\documentclass[journal]{IEEEtran}
\pdfoutput=1
\ifCLASSINFOpdf
\else
   \usepackage[dvips]{graphicx}
\fi
\usepackage{url}

\usepackage{graphicx}
\usepackage{bbding}
\usepackage{amsmath,amssymb}
\usepackage{booktabs}
\usepackage{cite}
\usepackage{subfigure}
\usepackage{float}
\usepackage{balance}
\begin{document}
\title{Deep Joint Source-Channel Coding for\\ Multi-Task Network}

\author{Mengyang Wang, Zhicong Zhang, Jiahui Li, \IEEEmembership{Member, IEEE}, \\Mengyao Ma, \IEEEmembership{Member, IEEE}, and Xiaopeng Fan, \IEEEmembership{Senior Member, IEEE}\vspace{-2mm}
\thanks{This work was supported in part by the National Science Foundation of China (NSFC) under grants 61972115 and 61872116. \textit{(Corresponding author: Xiaopeng Fan.)}}
\thanks{Mengyang Wang and Zhicong Zhang are with the School of Computer Science, Harbin Institute of Technology, Harbin 150001, China, and also with the Wireless Technology Lab, Huawei, Shenzhen 518129, China. Jiahui Li and Mengyao Ma are with the Wireless Technology Lab, Huawei, Shenzhen 518129, China (e-mail: lijiahui666@huawei.com; ma.mengyao@huawei.com). Xiaopeng Fan is with the School of Computer Science, Harbin Institute of Technology, Harbin 150001, China (e-mail: fxp@hit.edu.cn).}
}

\maketitle
\begin{abstract}
Multi-task learning (MTL) is an efficient way to improve the performance of related tasks by sharing knowledge. However, most existing MTL networks run on a single end and are not suitable for collaborative intelligence (CI) scenarios. In this work, we propose an MTL network with a deep joint source-channel coding (JSCC) framework, which allows operating under CI scenarios. We first propose a feature fusion based MTL network (FFMNet) for joint object detection and semantic segmentation. Compared with other MTL networks, FFMNet gets higher performance with fewer parameters. Then FFMNet is split into two parts, which run on a mobile device and an edge server respectively. The feature generated by the mobile device is transmitted through the wireless channel to the edge server. To reduce the transmission overhead of the intermediate feature, a deep JSCC network is designed. By combining two networks together, the whole model achieves 512$\times$ compression for the intermediate feature and a performance loss within 2\% on both tasks. At last, by training with noise, the FFMNet with JSCC is robust to various channel conditions and outperforms the separate source and channel coding scheme.
\end{abstract}

\begin{IEEEkeywords}
Collaborative intelligence, multi-task learning, deep joint source-channel coding
\end{IEEEkeywords}

\IEEEpeerreviewmaketitle

\section{Introduction}

\IEEEPARstart{W}{ith} the development of deep learning \cite{1}, the convolutional neural network (CNN) is playing an important role in computer vision tasks, like object detection\cite{2} and semantic segmentation\cite{3}. These tasks are usually studied as separate problems. However, preparing the model for each task individually (i.e., the so-called single-task learning, STL) is inefficient and storage-intensive. Therefore, multi-task learning (MTL)\cite{9} emerges as the times require. In MTL, related tasks are handled using one model, where common features are obtained by sharing a backbone, and different branches are used to solve different tasks. Compared with STL, MTL is more efficient and storage-saving. There are some excellent studies of joint object detection and semantic segmentation \cite{10,12,11}. However, BlitzNet \cite{10} and TripleNet \cite{12} have multiple branches, skipping connections, which makes the backbone complex. DSPNet \cite{11} has many layers and parameters, resulting in high model storage costs. None of these models are suitable for running on mobile devices.

To enable complex models running on mobile devices, a paradigm called collaborative intelligence (CI) \cite{13} has emerged. In CI, a deep model is split between the mobile device and the edge server, which balances the computational load between them. The mobile device extracts and transmits intermediate features from the input signal, and the edge server receives and processes these features. To reduce the transmission overhead, some feature compression methods have been proposed in \cite{14,15,16,17}. In \cite{14} and \cite{15}, HEVC-intra and HEVC-inter are used to compress features, respectively. PNG and JPEG are used to compress features in \cite{16} and \cite{17}. But these models do not take into account transmission errors due to channel noise or interference.

In traditional communication systems, source coding and channel coding are two separate steps. Nowadays, joint source-channel coding (JSCC) is known to outperform the separate approach in practical applications \cite{26}. Recently, the deep learning based joint source-channel coding method \cite{18} has been proposed, which uses an auto-encoder (AE) to compress and transmit images or features over wireless channels. In\cite{19} and\cite{20}, a CNN based AE is used to compress and transmit images. In\cite{21,22,23}, AE is used to compress the intermediate feature of a STL network. These deep models are trained with noise, so they are robust to channel interference.

To get an MTL network that is suitable for CI and robust to channel interference, in this letter, we propose an MTL network with deep JSCC for the intermediate feature. Firstly, an MTL network (FFMNet) is designed to handle object detection and semantic segmentation. Then FFMNet is split into mobile device part and edge server part. To reduce the transmission overhead, we propose a new AE based JSCC framework to compress and transmit the intermediate feature of FFMNet. Finally, the robust JSCC model for MTL is obtained through training the whole model with channel noise. To the best of the authors' knowledge, this is the first work to perform JSCC for MTL. Our main contributions are summarized as follows:
\begin{enumerate}[]
\item We propose a novel MTL network, FFMNet, for joint object detection and semantic segmentation. And FFMNet achieves better performance and has fewer parameters than other frameworks.
\item A JSCC network is designed to compress and transmit the intermediate feature of FFMNet, which achieves 512$\times$ compression with a performance loss under 2\% on both tasks.
\item The FFMNet with JSCC model is trained to achieve robust transmission over noisy channel. Simulation results show that the performance of the proposed model is much better than that of the separate coding scheme.
\end{enumerate}

The rest part of this paper is as follows. In Section \ref{sec:2}, we show the overall architecture, the details of the FFMNet and the JSCC model, and then introduce the loss functions and training strategy. The dataset and the performance of the FFMNet and the JSCC model are shown in Section \ref{sec:3}. Finally, the paper is concluded in Section \ref{sec:4}.

\section{Methodology}
\label{sec:2}
\subsection{Overall Network Architecture}
\begin{figure}[htbp]
\centerline{\includegraphics[width=\columnwidth]{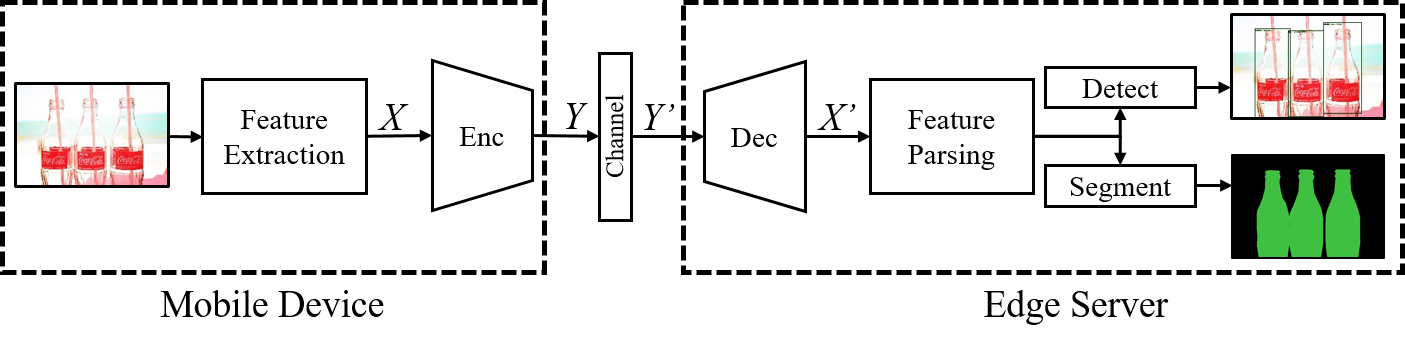}}
\caption{The overall architecture of the proposed network.}
\label{fig:1}
\end{figure}
The overall architecture of the proposed framework is shown in Fig. \ref{fig:1}. It contains a mobile device and an edge server. At the mobile device, the input image is first processed by a feature extraction module, which obtains deep intermediate feature $X$ containing enough information for the considered tasks. Since $X$ can be too large in size, transmitting it to the edge server requires significant communication resources. To reduce the communication overhead, an encoder compresses $X$ into a compact feature $Y$, and then sends $Y$ to the edge server through a noisy channel. At the edge server, noisy feature $Y^{\prime}$ is received and processed by a decoder to get the reconstruction $X^{\prime}$. Finally, the edge server parses $X^{\prime}$ using the feature parsing module to obtain the features to be shared by object detection and semantic segmentation.

The wireless channel between the mobile device and the edge server is modeled by an additive white Gaussian noise (AWGN) model. Given an input feature $z \in \mathbb R^{B}$ and an output feature $z^{\prime} \in \mathbb R^{B}$, the transfer function of AWGN channel is written as $z^{\prime}=z+n$, with $n \sim \mathcal{N}(0, \sigma^2)$. The $\sigma^2$ is the noise variance, which denotes the channel condition. Since the function is differentiable, the network can be trained end-to-end. Besides, to meet the average transmit power constraint of $P=1$, i.e. $\frac{1}{B} \sum\limits_{i=1}^{B} z_i^2 = P$, a power normalization layer is put at the end of the encoder. We evaluate the performance of the object detection and semantic segmentation for different channel signal-to-noise ratios (SNRs) defined as $\frac {P}{\sigma^2}$.

\subsection{Feature Fusion based Multi-task Network (FFMNet)}
\begin{figure}[htbp]
\centerline{\includegraphics[width=\columnwidth]{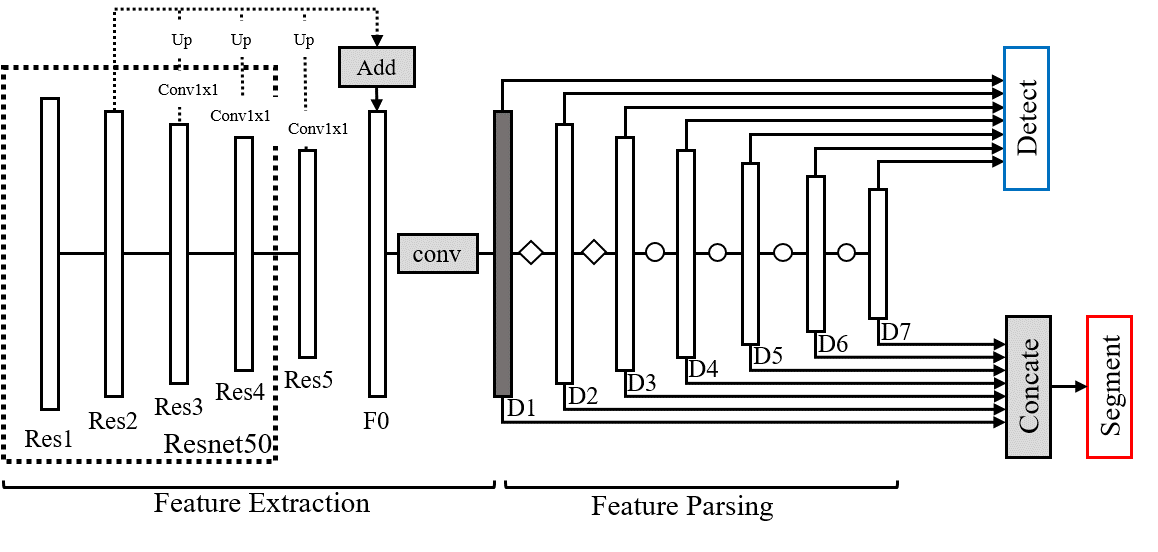}}
\caption{The architecture of FFMNet.}
\label{fig:2}
\end{figure}
Nowadays, MTL is getting more attention, and there are many excellent frameworks. The BlitzNet \cite{10} and TripleNet \cite{12} perform object detection and semantic segmentation, and they exhibit high performance on both tasks at a high speed. In BlitzNet, ResNet50 \cite{27} is used as the basic model, and the two branches perform their respective tasks by sharing a set containing multiscale features. There are many residual connections and ‘ResSkip’ blocks in the backbone to fuse the features of the front and back sides, which makes the backbone very complex. To get an MTL network that is lightweight and CI-suitable, we design a feature fusion based multi-task network FFMNet as shown in Fig. \ref{fig:2}. In FFMNet, ResNet50 is also used as the basic model and multi-scale features are used by the two tasks. The difference with BlitzNet is the way of getting the multi-scale features set from the basic model. According to Fig. \ref{fig:2}, FFMNet consists of three parts: feature extraction, feature parsing, and task branches.

In the feature extraction part, the Res$_1$ to Res$_4$ are generated by ResNet50 and Res$_5$ is an additional layer. Then the feature fusion technology similar to FSSD\cite{5} is applied to fuse the Res$_2$ to Res$_5$ into one feature F$_0$. The dimension of F$_0$ is (64, 64, 512) and is represented by $(height, width, channels)$. We first use convolutional layers with 1$\times$1 kernels to convert the channels of Res$_2$ to Res$_5$ into 512, then the bilinear upsampling method is used to convert the spatial dimensions of Res$_2$ to Res$_5$ into 64$\times$64. After that, the resized features are added together as F$_0$. At last, the final feature D$_1$ is  generated using a convolutional layer with 3$\times$3 kernels.

\begin{figure}[htbp]
\centerline{\includegraphics[width=0.7\columnwidth]{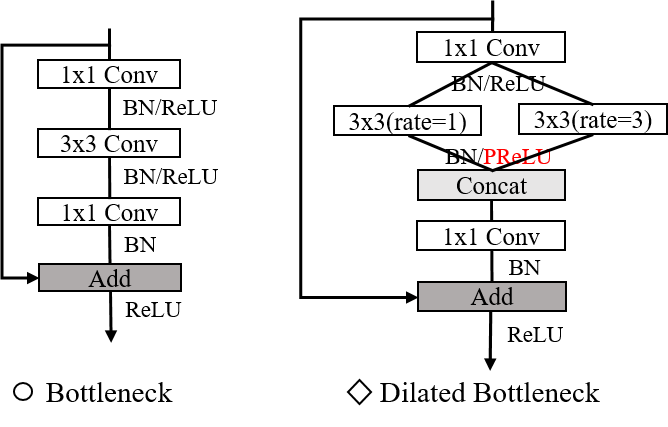}}
\caption{The Bottleneck and Dilated Bottleneck used in feature parsing part.}
\label{fig:3}
\end{figure}

In the feature parsing part, feature D$_1$ is parsed into a set of multi-scale features to perform the two tasks. The scales from feature D$_1$ to D$_7$ are (64, 64, 512), (32, 32, 512), ..., (1, 1, 512), respectively. The reason for using multi-scale features is that different scale features have different reception fields, which helps to identify objects of different scales \cite{4}. As shown in Fig. \ref{fig:3}, the `Bottleneck' of ResNet and `Dilated Bottleneck' are used to generate features with different scales, and the `Dilated Bottleneck' is formed based on `Bottleneck' and dilated convolutional layer. Since the sizes of D$_1$ and D$_2$ are much bigger than others, the `Dilated Bottleneck' with larger receptive field is employed. D$_2$ and D$_3$ are generated by the `Dilated Bottleneck'. The remaining small-scale features are generated by `Bottleneck'.

After getting the feature set, object detection head and semantic segmentation head share the features to perform their own tasks, and these heads are the same as those used in BlitzNet. In the detection branch, there are two convolutional layers with 3$\times$3 kernels for classification and location regression respectively for each size of features in the set. Then the non-maximum suppression (NMS) is employed as the post-processing method to eliminate redundant detection results. On the other side, the segmentation branch first uses upsampling method and 1$\times$1 convolutional layer to resize the features into the same size (64, 64, 64). Then the rescaled features are concatenated together in channel dimension. At last, a convolutional layer with 3$\times$3 kernels is to predict the classes of each pixel of the image.

\subsection{JSCC of Intermediate feature}
\begin{figure}[htbp]
\centerline{\includegraphics[width=\columnwidth]{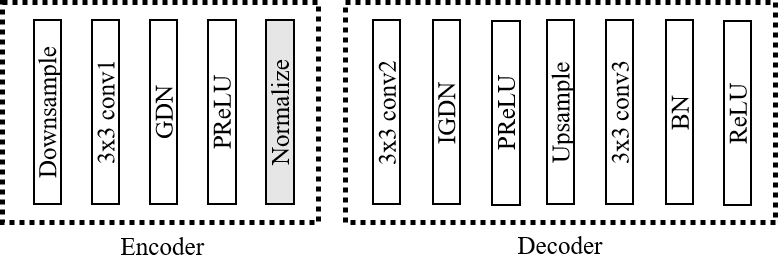}}
\caption{The proposed JSCC encoder and decoder.}
\label{fig:4}
\end{figure}

With the idea of CI and the structure of FFMNet introduced before, we split FFMNet into two parts and apply a JSCC network on the intermediate feature. We choose D$_1$ as the split point and propose an AE based JSCC network. The structure of the JSCC network is shown in Fig. \ref{fig:4}.

Inspired by \cite{22}, an asymmetric structure is designed to reduce computation on the device, so there are fewer layers in the encoder than in the decoder. In the encoder, the downsampling method is proposed to reduce the spatial size of the feature, and the number of feature channels is reduced by a convolutional layer with 3$\times$3 kernels. Then, the generalized divisive normalization (GDN)\cite{24} layer follows the convolutional layer. At the end of the encoder, we place a power normalization layer. In the decoder, two convolutional layers with 3$\times$3 kernels recover the channels gradually, and an upsampling layer between them is used to recover the spatial size. The inverse generalized divisive normalization (IGDN) layer performs the inverse operation of the GDN layer.

Compare to the AE in \cite{22}, the proposed JSCC network has some advantages: First, the downsampling layer can downsample the spatial size with stride 4, which can not be done with 3$\times$3 convolutional layers. Second, in the decoder, we use two convolutional layers to recover the channels gradually, which helps to reconstruct the feature. Third, the last activation function is ReLU, which makes the reconstructed feature in the same value range as D$_1$. 

By combining this network with the FFMNet, we finally get the JSCC architecture for MTL. 

\subsection{Training Strategy}
\label{sec:2-4}
When training FFMNet, the loss function $\mathcal L_{MTL}$ has two parts: the loss function of object detection $\mathcal L_{det}$ and semantic segmentation $\mathcal L_{seg}$. For detection, we use the same loss function in SSD\cite{4}, which includes a classification loss $\mathcal L_{cls}$ and a regression loss of locating the bounding boxes $\mathcal L_{bbox}$. So the loss function of detection is
\begin{equation}
\mathcal L_{det} = \mathcal L_{cls} + \mathcal L_{bbox}.
\end{equation}

For segmentation, the loss is the cross-entropy between predicted and target class distribution. We add the loss functions of two branches as the final loss function:
\begin{equation}
\mathcal L_{MTL} = \mathcal L_{det} + \mathcal L_{seg}.
\end{equation}

To fully train the network, we propose a three-step training strategy: Firstly $\mathcal L_{MTL}$ is employed to train the FFMNet. Then we attach the JSCC network at the split point, the sum of the L1 loss between $X$ and $X^{\prime}$ and the $\mathcal L_{MTL}$ is the loss function to train the JSCC part, and other parameters are fixed. Finally, we use $\mathcal L_{MTL}$ to train the parameters of the whole network end-to-end.

To make the whole system robust to the channel noise, the network will be trained at different SNRs by changing the $\sigma^2$.
\section{Experiments}
\label{sec:3}
\subsection{Dataset}
In the experiments, we use part of the Open Images Dataset\cite{25} as training and testing data. The training set contains 178847 images and testing set contains 9903 images. All the images are labeled with the annotations of object detection and semantic segmentation, and there are 117 object categories in the annotation.

\subsection{Performance for FFMNet}
At the first step of the training strategy, FFMNet is optimized by the Adam algorithm with a mini-batch size of 32 images. The iteration number is 500K, and the initial learning rate is set to $10^{-4}$ and decreased twice by factor 10 at iteration 250K and 450K.
\begin{table}[H]
\renewcommand{\arraystretch}{1.3}
     \caption{Comparison Between MTL Networks}
	\label{tab:table1}
     \centering
     \scalebox{0.9}{
	\begin{tabular}{c|cccc}
		\toprule 
		Network   & mAP(\%)         & mIoU(\%)             & Param          \\
		\midrule
		FFMNet    & 40.8            & \textbf{44.6}    & \textbf{63.2M} \\
		BlitzNet  & 40.1            & 44.1             & 87.8M          \\
           TripleNet  & \textbf{41.4}  & 44.0             & 137.3M          \\
		\bottomrule
	\end{tabular}
     }
\end{table}
According to the training settings, FFMNet is well trained. In Table \ref{tab:table1}, we compare the performance on two tasks and the number of parameters for the entire network of FFMNet, BlitzNet and TripleNet. The metric for evaluating detection performance is the mean average precision (mAP) and the quality of predicted segmentation masks is measured with mean intersection over union (mIoU). It is shown that FFMNet performs better than BlitzNet while has fewer parameters. Compared with TripleNet, FFMNet performs slightly worse on detection but better on segmentation. Besides, the number of parameters is reduced by more than 50\%, making FFMNet more suitable for CI scenarios than the baseline methods.

\begin{table}
	\centering
	\caption{Comparison Between MTL And STL}
     \scalebox{0.9}{
	\begin{tabular}{c|cccc}
		\toprule
		Network   & Det         & Seg       & mAP(\%)        & mIoU(\%)          \\
		\midrule
		FFMNet    & \checkmark  &\checkmark & \textbf{40.8}  & \textbf{44.6}    \\
		FFMNet    & \checkmark  &           & 40.3           & -                \\
		FFMNet    &             &\checkmark & -              & 40.4             \\
		\bottomrule
	\end{tabular}}
	\label{tab:table2}
\end{table}

To verify whether MTL outperforms STL, we train two STL networks respectively, and the results are shown in Table \ref{tab:table2}. As can be seen from Table \ref{tab:table2}, joint detection and segmentation improves detection and segmentation performance by 0.5\% and 4.2\%, respectively, indicating that the two tasks are mutually beneficial in FFMNet.

\subsection{Performance for JSCC model}

The proposed JSCC network is trained by the strategy described in Section \ref{sec:2-4}, on the basis of FFMNet.

In the second step of the training strategy, we train the encoder and decoder for 100K iterations, the learning rate is set to $10^{-4}$ and decreased by 10 at iteration 50K. In the third step, the whole network is trained end-to-end for 200K iterations, the learning rate is set to $10^{-5}$ and decreased by 10 at iteration 100K and 150K. 

To explore the compression capability of the JSCC model, a group of JSCC models with different compression ratios for the intermediate feature of FFMNet are trained under the noiseless condition. The performance of different JSCC models is shown in Table \ref{tab:table3}, where the original feature dimension is (H, W, C). With performance degradation threshold set to 2\% on both tasks, the JSCC model achieves up to 512$\times$ compression of the intermediate feature.

In addition to the compression capability, robustness also needs to be considered in practice. To find the best training SNR (i.e., SNR$_{train}$) for JSCC, multiple networks are trained with channel noise under different SNRs, i.e., SNR$_{train}$ = 0, 5, and 10 dB, where the compression ratio is fixed to 512$\times$.

\begin{figure}
\subfigure[]{                
\begin{minipage}{\linewidth}
\centering                        
\includegraphics[width=0.75\columnwidth]{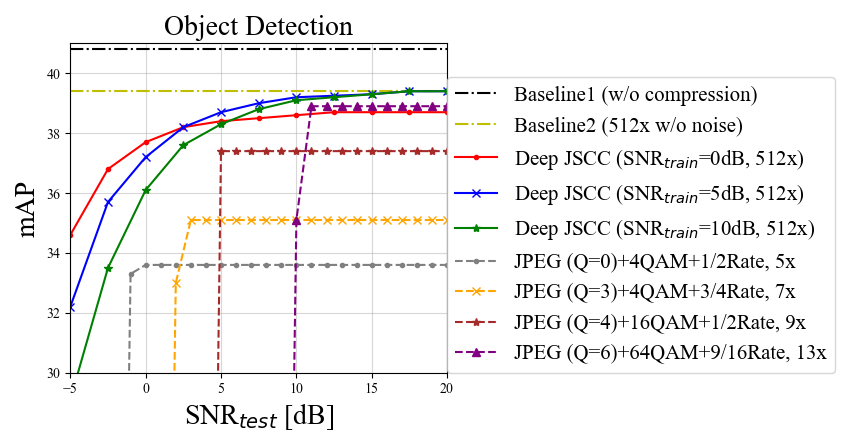}
\end{minipage}
}
\quad            
\subfigure[]{                
\begin{minipage}{\linewidth}
\centering                             
\includegraphics[width=0.75\columnwidth]{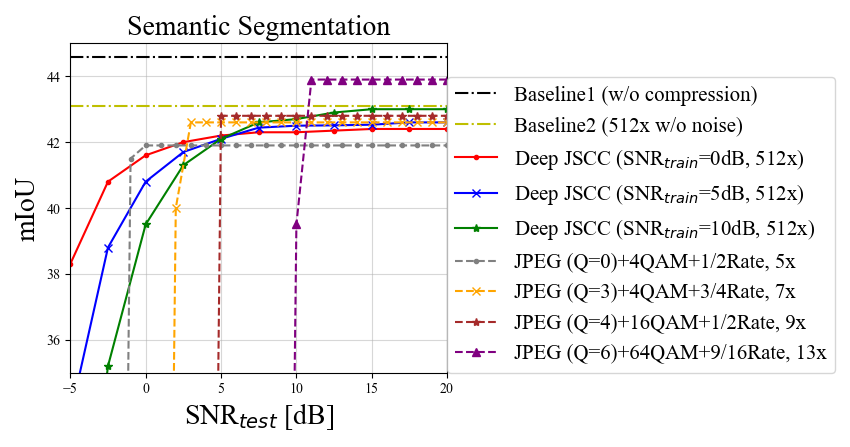}
\end{minipage}
}
\caption{Performance of detection and segmentation with respect to testing SNR (i.e., SNR$_{test}$)}
\label{fig:5}
\end{figure}

We plot the mAP of detection and the mIoU of segmentation under different test channel SNRs in Fig. \ref{fig:5}. In the figure, there are two baselines. The first baseline is the performance of the FFMNet without compression. The second baseline is the performance of the JSCC model with 512$\times$ compression, which is trained and tested without channel noise. Besides, the performance of the proposed deep JSCC for the intermediate feature is compared with separate schemes that use JPEG for feature compression followed by practical channel coding and modulation. 

For separate scheme, we first perform 8-bit uniform quantization on feature. Then, JPEG with different quality parameters is used to compress the quantized feature. At last, channel coding and modulation are employed. We take four combinations of source and channel coding to get the similar performance of the proposed JSCC method. It is obvious that the JSCC method outperforms the separate scheme in Fig. \ref{fig:5}. On one hand, with similar performance, the maximum compression ratio of the separate scheme is 13$\times$ while JSCC is 512$\times$. On the other hand, the proposed JSCC model does not suffer the `cliff effect' in separate method.

For JSCC, with the decrease of SNR$_{train}$, although the two task branches become more robust to the noise at low SNR values, the performance of both tasks degrades a little at high SNR. To get the trade-off between performance and robustness, the model trained with SNR$_{train}$ = 5 dB behaves well, so we select it as the final JSCC model for MTL. The final model is CI-suitable and robust to channel inference.

\begin{table}
	\centering
	\caption{Task Performance vs. Compression Ratio}
     \scalebox{0.9}{
	\begin{tabular}{c|c|c|c}
		\toprule
		Feature Size   &Compression Ratio       & mAP(\%)             &mIoU(\%)   \\
		\midrule
           (H,W,C)        & -                      & 40.8           & 44.6 \\
           (H/2,W/2,C/32) &128$\times$             & 40.2 (-0.6)    & 43.6 (-1.0) \\ 
           (H/2,W/2,C/64) &256$\times$             & 39.7 (-1.1)    & 43.1 (-1.5) \\
           (H/4,W/4,C/32) &512$\times$             & 39.4 (-1.4)    & 43.1 (-1.5) \\
           (H/4,W/4,C/64) &1024$\times$            & 38.8 (-2.0)    & 42.4 (-2.2) \\
		\bottomrule
	\end{tabular}}
	\label{tab:table3}
\end{table}

\section{Conclusion}
\label{sec:4}
In this letter, we study the MTL network which is CI-suitable, lightweight, and robust to channel interference. We first propose an MTL network named FFMNet, which performs object detection and semantic segmentation in the meantime. Results show that FFMNet gets higher performance and has fewer parameters than the baseline methods. Then we split the FFMNet into two parts and propose a JSCC scheme for efficient feature compression and robust feature transmission over the AWGN channel. The JSCC scheme achieves 512$\times$ compression for the intermediate feature. Besides, the model trained with SNR$_{train}$ = 5 dB exhibits robustness over a wide range of SNR$_{test}$ and outperforms the separate method a lot.

\balance
\bibliographystyle{IEEEtran}
\bibliography{refs}

\begin{thebibliography}{10}
\providecommand{\url}[1]{#1}
\csname url@samestyle\endcsname
\providecommand{\newblock}{\relax}
\providecommand{\bibinfo}[2]{#2}
\providecommand{\BIBentrySTDinterwordspacing}{\spaceskip=0pt\relax}
\providecommand{\BIBentryALTinterwordstretchfactor}{4}
\providecommand{\BIBentryALTinterwordspacing}{\spaceskip=\fontdimen2\font plus
\BIBentryALTinterwordstretchfactor\fontdimen3\font minus
  \fontdimen4\font\relax}
\providecommand{\BIBforeignlanguage}[2]{{%
\expandafter\ifx\csname l@#1\endcsname\relax
\typeout{** WARNING: IEEEtran.bst: No hyphenation pattern has been}%
\typeout{** loaded for the language `#1'. Using the pattern for}%
\typeout{** the default language instead.}%
\else
\language=\csname l@#1\endcsname
\fi
#2}}
\providecommand{\BIBdecl}{\relax}
\BIBdecl

\bibitem{1}
I.~Goodfellow, Y.~Bengio, A.~Courville, and Y.~Bengio, \emph{Deep
  learning}.\hskip 1em plus 0.5em minus 0.4em\relax MIT press Cambridge, 2016,
  vol.~1, no.~2.

\bibitem{2}
I.~Park and S.~Kim, ``Performance indicator survey for object detection,'' in
  \emph{2020 20th International Conference on Control, Automation and Systems
  (ICCAS)}.\hskip 1em plus 0.5em minus 0.4em\relax IEEE, 2020, pp. 284--288.

\bibitem{3}
F.~Lateef and Y.~Ruichek, ``Survey on semantic segmentation using deep learning
  techniques,'' \emph{Neurocomputing}, vol. 338, pp. 321--348, 2019.

\bibitem{9}
Y.~Zhang and Q.~Yang, ``An overview of multi-task learning,'' \emph{National
  Science Review}, vol.~5, no.~1, pp. 30--43, 2018.

\bibitem{10}
N.~Dvornik, K.~Shmelkov, J.~Mairal, and C.~Schmid, ``Blitznet: A real-time deep
  network for scene understanding,'' in \emph{Proceedings of the IEEE
  international conference on computer vision}, 2017, pp. 4154--4162.

\bibitem{12}
J.~Cao, Y.~Pang, and X.~Li, ``Triply supervised decoder networks for joint
  detection and segmentation,'' in \emph{Proceedings of the IEEE/CVF Conference
  on Computer Vision and Pattern Recognition}, 2019, pp. 7392--7401.

\bibitem{11}
L.~Chen, Z.~Yang, J.~Ma, and Z.~Luo, ``Driving scene perception network:
  Real-time joint detection, depth estimation and semantic segmentation,'' in
  \emph{2018 IEEE Winter Conference on Applications of Computer Vision
  (WACV)}.\hskip 1em plus 0.5em minus 0.4em\relax IEEE, 2018, pp. 1283--1291.

\bibitem{13}
Y.~Kang, J.~Hauswald, C.~Gao, A.~Rovinski, T.~Mudge, J.~Mars, and L.~Tang,
  ``Neurosurgeon: Collaborative intelligence between the cloud and mobile
  edge,'' \emph{ACM SIGARCH Computer Architecture News}, vol.~45, no.~1, pp.
  615--629, 2017.

\bibitem{14}
H.~Choi and I.~V. Baji{\'c}, ``Deep feature compression for collaborative
  object detection,'' in \emph{2018 25th IEEE International Conference on Image
  Processing (ICIP)}.\hskip 1em plus 0.5em minus 0.4em\relax IEEE, 2018, pp.
  3743--3747.

\bibitem{15}
H.~Choi and I.~V. Baji{\'c}, ``Near-lossless deep feature compression for
  collaborative intelligence,'' in \emph{2018 IEEE 20th International Workshop
  on Multimedia Signal Processing (MMSP)}.\hskip 1em plus 0.5em minus
  0.4em\relax IEEE, 2018, pp. 1--6.

\bibitem{16}
S.~R. Alvar and I.~V. Baji{\'c}, ``Multi-task learning with compressible
  features for collaborative intelligence,'' in \emph{2019 IEEE International
  Conference on Image Processing (ICIP)}.\hskip 1em plus 0.5em minus
  0.4em\relax IEEE, 2019, pp. 1705--1709.

\bibitem{17}
S.~R. Alvar and I.~V. Baji{\'c}, ``Bit allocation for multi-task collaborative
  intelligence,'' in \emph{ICASSP 2020-2020 IEEE International Conference on
  Acoustics, Speech and Signal Processing (ICASSP)}.\hskip 1em plus 0.5em minus
  0.4em\relax IEEE, 2020, pp. 4342--4346.

\bibitem{26}
F.~Zhai, Y.~Eisenberg, and A.~K. Katsaggelos, ``Joint source-channel coding for
  video communications,'' \emph{Handbook of Image and Video Processing}, pp.
  1065--1082, 2005.

\bibitem{18}
L.~Rongwei, W.~Lenan, and G.~Dongliang, ``Joint source/channel coding
  modulation based on bp neural networks,'' in \emph{International Conference
  on Neural Networks and Signal Processing, 2003. Proceedings of the 2003},
  vol.~1.\hskip 1em plus 0.5em minus 0.4em\relax IEEE, 2003, pp. 156--159.

\bibitem{19}
E.~Bourtsoulatze, D.~B. Kurka, and D.~G{\"u}nd{\"u}z, ``Deep joint
  source-channel coding for wireless image transmission,'' \emph{IEEE
  Transactions on Cognitive Communications and Networking}, vol.~5, no.~3, pp.
  567--579, 2019.

\bibitem{20}
D.~Burth~Kurka and D.~G{\"u}nd{\"u}z, ``Joint source-channel coding of images
  with (not very) deep learning,'' in \emph{International Zurich Seminar on
  Information and Communication (IZS 2020). Proceedings}.\hskip 1em plus 0.5em
  minus 0.4em\relax ETH Zurich, 2020, pp. 90--94.

\bibitem{21}
J.~Shao and J.~Zhang, ``Bottlenet++: An end-to-end approach for feature
  compression in device-edge co-inference systems,'' in \emph{2020 IEEE
  International Conference on Communications Workshops (ICC Workshops)}.\hskip
  1em plus 0.5em minus 0.4em\relax IEEE, 2020, pp. 1--6.

\bibitem{22}
M.~Jankowski, D.~Gunduz, and K.~Mikolajczyk, ``Deep joint
  transmission-recognition for power-constrained iot devices,'' \emph{arXiv
  preprint arXiv:2003.02027}, 2020.

\bibitem{23}
M.~Jankowski, D.~G{\"u}nd{\"u}z, and K.~Mikolajczyk, ``Deep joint
  source-channel coding for wireless image retrieval,'' in \emph{ICASSP
  2020-2020 IEEE International Conference on Acoustics, Speech and Signal
  Processing (ICASSP)}.\hskip 1em plus 0.5em minus 0.4em\relax IEEE, 2020, pp.
  5070--5074.

\bibitem{27}
K.~He, X.~Zhang, S.~Ren, and J.~Sun, ``Deep residual learning for image
  recognition,'' in \emph{Proceedings of the IEEE Conference on Computer Vision
  and Pattern Recognition (CVPR)}, June 2016.

\bibitem{5}
Z.~Li and F.~Zhou, ``Fssd: feature fusion single shot multibox detector,''
  \emph{arXiv preprint arXiv:1712.00960}, 2017.

\bibitem{4}
W.~Liu, D.~Anguelov, D.~Erhan, C.~Szegedy, S.~Reed, C.-Y. Fu, and A.~C. Berg,
  ``Ssd: Single shot multibox detector,'' in \emph{European conference on
  computer vision}.\hskip 1em plus 0.5em minus 0.4em\relax Springer, 2016, pp.
  21--37.

\bibitem{24}
J.~Ball{\'e}, V.~Laparra, and E.~P. Simoncelli, ``Density modeling of images
  using a generalized normalization transformation,'' \emph{arXiv preprint
  arXiv:1511.06281}, 2015.

\bibitem{25}
A.~Kuznetsova, H.~Rom, N.~Alldrin, J.~Uijlings, I.~Krasin, J.~Pont-Tuset,
  S.~Kamali, S.~Popov, M.~Malloci, A.~Kolesnikov \emph{et~al.}, ``The open
  images dataset v4,'' \emph{International Journal of Computer Vision}, pp.
  1--26, 2020.

\end{thebibliography}
\end{document}